\pdfoutput=1

\documentclass[11pt]{article}

\usepackage{emnlp2021-latex/emnlp2021}

\usepackage{times}
\usepackage{latexsym}
\usepackage{pifont}
\usepackage{amssymb}

\usepackage[T1]{fontenc}

\usepackage[utf8]{inputenc}

\usepackage{microtype}

\usepackage{style/style}

\title{Written Justifications are Key to Aggregate Crowdsourced Forecasts}

\author{Saketh Kotamraju \\
  Heritage High School\\
  \texttt{saketh.kotamraju@gmail.com} \\\And
  Eduardo Blanco \\
  Arizona State University \\
  \texttt{eduardo.blanco@asu.edu} \\}

\begin{document}
\maketitle
\begin{abstract}
This paper demonstrates that aggregating crowdsourced forecasts benefits from modeling the written justifications provided by forecasters.
Our experiments show that the majority and weighted vote baselines are competitive,
and that the written justifications are beneficial to call a question throughout its life except in the last quarter.
We also conduct an error analysis shedding light into the characteristics that make a justification unreliable. \\
\end{abstract}

\section{Introduction}
\label{s:introduction}

The wisdom of the crowd refers to the idea that aggregating information collected from many nonexperts
often yields good answers to questions---as close to the truth or even better than asking an expert.
Perhaps the best known example is by~\newcite{Galton1907Voxpopuli},
who observed that the median estimate of the weight of an ox (out of 800 country fair attendees) was within 1\% of the truth.
There is a lot of support for the idea, although it is well know that it is not foolproof~\cite{10.5555/1095645}.
\newcite{madness} presents historical examples where crowds behaved irrationally,
and more recently, world chess champion Gary Kasparov beat the crowd playing chess~\cite{marko1999kasparov}.

In this day and age, the benefits of the crowd are commonplace.
Wikipedia is written by volunteers,
and community question answering has received the attention of researchers~\cite{
adamic2008knowledge,
wang2013wisdom}.
When aggregating information collected from crowds,
it may be important to know whether judgments were collected independently of each other.
If they were not,
crowd psychology~\cite{doi:https://doi.org/10.1002/9780470998458.ch8}
and the power of persuasion~\cite{o2015persuasion}
can bias individual judgments and eliminate the wisdom of the crowd.

\begin{figure}
\small
\centering
\begin{tabular}{@{}p{\columnwidth}@{}}
\toprule
Question: Will there be a new prime minister of Italy before 1 September 2021? \\
Start date: 1/28/2021, closing date: 2/13/2021 \\ \midrule

Forecast 1: 100\% yes, 0\% no \\
Justification: \emph{Actually the media talk about potential candidates [link] the Crowd is 98\% Yes} \\ \midrule

Forecast 2: 99\% yes, 1\% no \\
Justification: \emph{With a substantial majority now backing Draghi (who in turns seems to be an obvious EU favourite which brings better prospects for bail out funding) this seems to be a virtual certainty at this stage. [link] 
Thanks [user] for digging up the parliamentary numbers! [link] [link]} \\ \bottomrule
\end{tabular}

\caption{Question and forecasts submitted by the crowd.
  Justifications provide information about the credibility of the forecast.
  The first justification is weak and refers to the current opinion of the crowd;
  the second justification is strong and provides links to support the claims.}
\label{f:motivation}
\end{figure}

In this paper, we work with forecasts about
questions across the political, economic, and social spectrum.
Each forecast consists of a prediction estimating the likelihood of some event and a written justification explaining the prediction.
As Figure~\ref{f:motivation} shows,
forecasts with the same predictions may come with
weaker or stronger justifications that affect the credibility of the predictions.
For example, the first justification refers to an external source without justifying why,
and it appears to rely on the current opinion of the crowd.
On the other hand,
the second justification provides specific facts from external resources and previous forecasters.

We move to a discussion of important terminology. We define a \emph{question} as a sentence that elicits information
(e.g., `Will legislation raising the US federal minimum wage become law before 20 August 2021?').
Questions have an \emph{opening} and \emph{closing day},
and the days in between are the \emph{life of the question}.
\emph{Forecasters} are people who submit a forecast.
A \emph{forecast} consists of a prediction and a justification.
The \emph{prediction} is a number indicating the chances that something will happen.
Following with the question above, a prediction could be `70\% chance' (of the legislation becoming law before 20 August 2021).
A \emph{justification} is the text forecasters submit in support of their predictions~(see examples in Figure~\ref{f:motivation} and Section~\ref{s:qualitative_analysis}). We use the phrase \emph{call a question} to refer to the problem we work with:
make a final prediction after aggregating individual forecasts.
We call questions each day throughout their life  using two strategies:
forecasts submitted in the given day~(\emph{daily})
and
the last forecast submitted by each forecaster~(\emph{active}).
Note that in this paper we use \emph{prediction} to refer to the submission by a forecaster,
not the output of a machine learning model.

Inspired by previous work
on identifying and cultivating better forecasters~\cite{mellers2015identifying},
and analyzing written justifications to estimate the quality of a single forecast~\cite{schwartz-etal-2017-assessing}
or
all forecasts by a forecaster~\cite{zong-etal-2020-measuring},
we experiment with the problem of automatically calling a question through its life based on the available forecasts in each day.
The main contributions of this paper are empirical results answering the following research questions: 
\begin{compactitem}
\item When calling a question on a particular day, is it worth taking into account forecasts submitted in previous days? (it is);
\item Does calling a question benefit from taking into account the question and the justifications submitted with the forecasts? (it does);
\item Is it easier to call a question towards the end of its life? (it is); and
\item Is it true that the worse the crowd predictions the more useful the justifications? (it is).
\end{compactitem}

In addition, we also present an analysis of the justifications submitted with correct and wrong forecasts to shed light into which characteristics make a justification more and less credible.
\section{Previous Work}
\label{s:previous_work}

\begin{figure*}[t!]
\centering
\begin{tabular}{@{}c@{}}

\includegraphics[width=\textwidth]{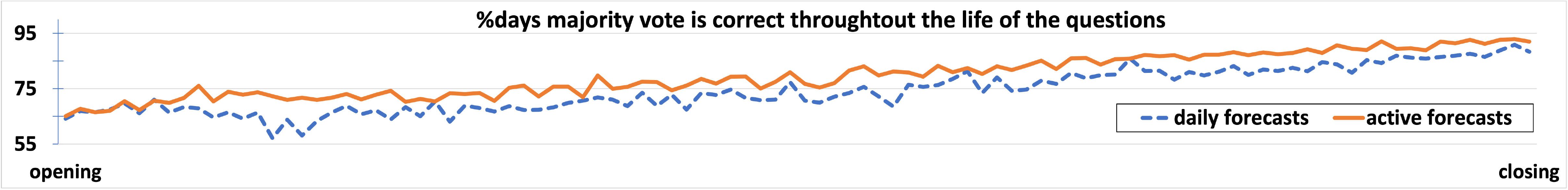} \\
\includegraphics[width=\textwidth]{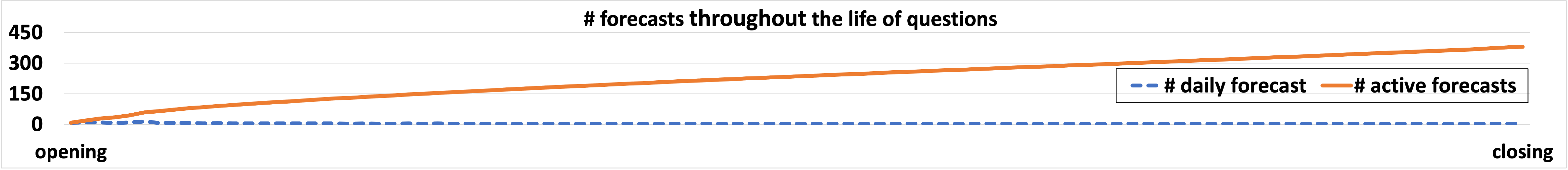}\\
\end{tabular}
\caption{Average number of daily and active forecasts available per question (bottom) and
average number of questions the majority forecast gets correct (top) over the life of the question (x-axis).
There is a tiny peak of forecasts submitted soon after a question is published and then a roughly uniform amount through the life of the question.
The majority of the forecasts, unsurprisingly, is less reliable towards the first half of the life of a question.
}
\label{f:dataset_summary}
\end{figure*}


The language people use is indicative of several attributes.
Previous work includes both predictive models (input: language samples, output: some attribute about the author)
and models that yield useful insights (input: language samples and attributes of the authors, output: differentiating language features depending on the attributes).
Among many others, previous research has studied gender and age~\cite{
li-etal-2016-semi,
nguyen-etal-2014-gender,
10.1145/2065023.2065035},
political ideology~\cite{
iyyer-etal-2014-political,
preotiuc-pietro-etal-2017-beyond},
health outcomes~\cite{schneuwly-etal-2019-correlating},
and personality traits~\cite{10.1371/journal.pone.0073791}.
In this paper, we do not profile forecasters.
Instead, we build models to call questions
based on forecasts by the crowd without knowledge of who submitted what.

Previous research has also studied the language people use to communicate depending on
the relationship between the parties.
For example, the language people use when they are in positions of power (e.g., more seniority)
has been studied in
social networks~\cite{bramsen-etal-2011-extracting},
online communities~\cite{10.1145/2187836.2187931},
and corporate emails~\cite{prabhakaran-rambow-2014-predicting}.
Similarly, \newcite{rashid-blanco-2018-characterizing}
study how language provides clues about the interactions and relationships between people.
Regarding language form and functions,
prior research has analyzed
politeness~\cite{danescu-niculescu-mizil-etal-2013-computational},
empathy~\cite{sharma-etal-2020-computational},
advice~\cite{govindarajan-etal-2020-help},
condolences~\cite{zhou-jurgens-2020-condolence}
usefulness~\cite{momeni2013properties},
and
deception~\cite{soldner-etal-2019-box}.
More related to the problem we work with,
\newcite{maki-etal-2017-roles} analyze the influence of Wikipedia editors,
and 
\newcite{katerenchuk2016hierarchy} study influence levels in online communities.
Persuasion has also been studied from a computational perspective~\cite{
wei-etal-2016-post,
yang-etal-2019-lets},
including dialogue systems~\cite{wang-etal-2019-persuasion}.
The work presented here complements these works.
We are interested in identifying credible justifications in order to aggregate crowdsourced forecasts,
and we do so without explicitly targeting any of the above characteristics.

Within computational linguistics, the previous task that is perhaps the closest to our goal is argumentation:
a good justification for a forecast is arguably a good supporting argument.
Previous work includes identifying argument components such as 
claims, premises, backings, rebuttals, and refutations~\cite{habernal-gurevych-2017-argumentation},
and mining supporting and opposing arguments for a claim~\cite{stab-etal-2018-cross}.
Notwithstanding these works,
we found that crowdsourced justifications rarely fall into these argumentation frameworks
despite the former are useful to aggregate forecasts.


Finally, there are a few works on forecasting that use the same or very similar corpora than we do.
From a psychology perspective,
\newcite{mellers2015identifying} present strategies
to improve forecasting accuracy (using top forecasters, i.e.,  superforecasters) and analyze 
the characteristics of superforecaster performance, which can be used for cultivating better forecasters.
\newcite{mellers2014psychological} discuss explanations of what makes forecasters better.
These works aim at identifying superforecasters and do not take into account the written justifications.
Unlike them, we build models to call questions without using any information about forecasters.
Within computational linguistics, 
\newcite{schwartz-etal-2017-assessing} assess the 
language of quality justifications (rating, benefit, and influence). 
\newcite{zong-etal-2020-measuring} is perhaps the closest experiment to ours. They build models to predict forecaster skill using the text justifications of forecasts from Good Judgment Open data, and they also use another dataset, Company Earnings Reports, to individually predict which forecasts are more likely to be correct predictions. 
Unlike us, none of these works aim at calling the question throughout its life.

\section{Dataset}
\label{s:terminology_dataset}


We work with data from the Good Judgment Open,\footnote{https://www.gjopen.com/} a website where questions are posted
and people submit forecasts.
Questions are about geopolitics
and include topics such as domestic and international politics, the economy, and social issues.
We collected all binary questions along with all their forecasts including a prediction and a justification.
In total, the dataset we work with contains 441 questions and 96,664 forecasts submitted in 32,708 days.
This is almost twice the amount of forecasts considered by \newcite{zong-etal-2020-measuring}.
Since our goal is to call questions throughout their life,
we work with all forecasts with written justifications regardless
of length,
how many forecasts have been submitted by the same forecaster, etc.
Additionally, our framework preserves privacy as we do not use any information about the forecaster.

The bottom plot in Figure \ref{f:dataset_summary} shows the average number of daily and active forecasts over the life of all questions.
There is roughly a uniform number of forecasts submitted each day,
thus the amount of active forecasts increases linearly over the life of the question.
The majority baseline with both daily and active forecasts submitted in the previous 10 days is quite accurate,
especially towards the closing date of questions.
The experiments presented in this paper aim at calling questions throughout their life.
As we shall see, models to automatically 
call questions benefit from taking into account justifications 
during the first three quarters of the life of a question.



\begin{table}
\small
\centering
\newcommand\cw{\hspace{.12in}}
\begin{tabular}{l r@{\cw}r@{\cw}r@{\cw}r@{\cw}r@{\hspace{.28in}}r}
\toprule
& Min & Q1 & Q2 & Q3 & Max & Mean \\ \midrule
\#tokens & 8 & 16 & 20 & 28 &48 & 21.94\\
\#entities & 0 & 2 & 3 & 5 & 11 & 3.47 \\
\#verbs & 0 & 2 & 2 & 3 & 6 & 2.26 \\
\#days open & 2 & 24 & 59 & 98 & 475 & 74.16\\
\bottomrule
\end{tabular}

\caption{Analysis of the questions from our dataset.
Most questions are relatively long, contain two or more named entities, and are open for over one month.}
\label{t:q_analysis}
\end{table}

\begin{figure}

\begin{tabular}{@{}c@{}c@{}}
\includegraphics[width=1.5in]{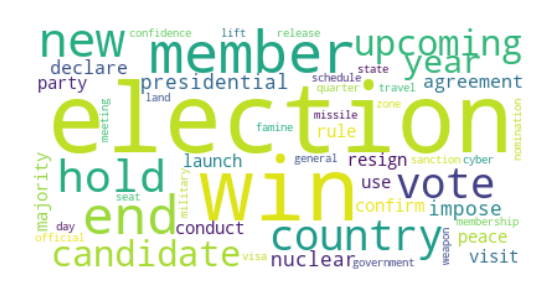}
&
\includegraphics[width=1.5in]{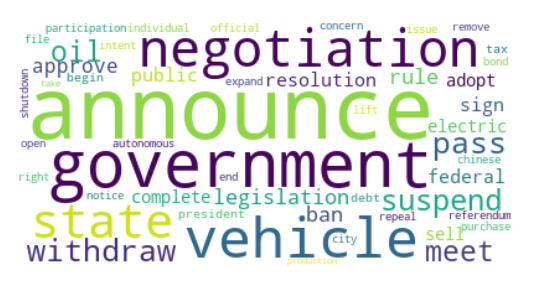} \\ \addlinespace

\multicolumn{2}{c}{\includegraphics[width=1.5in, height=0.75in]{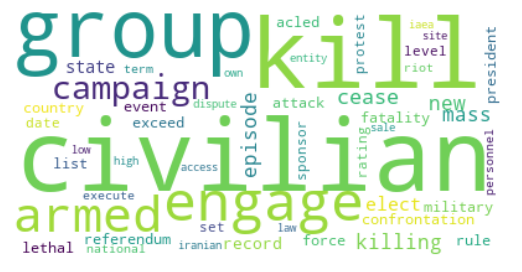}}\\
\end{tabular}

\caption{Topics obtained with LDA topic modeling in the 441 questions in our corpus.
The topics roughly correspond to (clockwise from top left)
(a)~elections, (b)~government actions, and (c)~war and violent events.}
\label{f:q_topics}
\end{figure}

\paragraph{Analyzing the Questions}
Table \ref{t:q_analysis} shows a basic analysis of the questions in our dataset.
The majority of questions have over 16 tokens and several entities;
the most common are \emph{geopolitical}, \emph{person} and \emph{date} entities.
Regarding the life of questions, we observe that half are open for almost two months, and 75\% for over three weeks.

Figure~\ref{f:q_topics} shows the LDA topics \cite{blei2003latent} obtained with gensim \cite{rehurek_lrec}.
We observe three main topics:
elections (voting, winners, candidate, etc.),
government actions (negotiations, announcements, meetings, passing (a law), etc.),
and
wars and violent crimes (groups killing, civilian (casualties), arms, etc.).
While not shown in the LDA topics,
the questions cover both domestic and international events in these topics.
 


\begin{table}
\small
\centering
\newcommand\cw{\hspace{.08in}}
\begin{tabular}{l r@{\cw}r@{\cw}r@{\cw}r@{\cw}r@{\cw}r}
\toprule
& Min & Q1 & Q2 & Q3 & Max\\ \midrule
\#sentences &  1 & 1 & 1 & 3 & 56 \\
\#tokens    &  1 & 10 & 23 & 47 & 1295 \\
\#entities  &  0 & 0 & 2 & 4 & 154 \\ \midrule

\#verbs     & 0 & 1 & 3 & 6 & 174 \\
\#adverbs   & 0 & 0 & 1 & 3 & 63 \\
\#adjectives & 0 & 0 & 2 & 4 & 91 \\ \midrule

\#negation  & 0 & 0 & 1 & 3 & 69  \\
Sentiment   & -2.54 & 0 & 0 & 0.20 & 6.50  \\ \midrule

Readability    \\
~~~Flesch     & -49.68 & 50.33 & 65.76 & 80.62 & 121.22 \\
~~~Dale-Chall & 0.05 & 6.72 & 7.95 & 9.20 & 19.77 \\
\bottomrule
\end{tabular}

\caption{Analysis of the 96,664 written justifications submitted by forecasters in our dataset.
The readability scores indicate that most justifications are easily understood by high school students (11th or 12th grade),
although a substantial amount (>25\%) require a college education (Flesch under 50 or Dale-Chall over 9.0).}
\label{t:j_analysis}
\end{table}

\paragraph{Analyzing the Justifications}

Table \ref{t:j_analysis} presents basic analysis of the 96,664 forecasts justification in our dataset.
The median length is short~(1~sentence and 23~tokens),
and justifications mention named entities less often than questions~(Table~\ref{t:q_analysis}).
We check whether justifications have negations using the cues annotated in ConanDoyle-neg~\cite{morante-daelemans-2012-conandoyle}.
Surprisingly, half of the justifications have one negation, and 25\% have three or more.
This indicates that forecasters sometimes rely on what may \emph{not} happen (or has \emph{not} happened) to make predictions about the future (questions do not have negations).
We also look at the sentiment polarity of justifications using TextBlob~\cite{textblob}.
The majority of justifications are neutral (polarity close to 0).
In terms of readability,
we compute the
Flesch~\cite{flesch1948new}
and
Dale-Chall~\cite{dale1948formula} scores.
Both scores indicate that around a quarter of justifications require a college education to be understood.

In terms of verbs and nouns,
we analyze them using the WordNet lexical files~\cite{miller1995wordnet}.
The most common verb classes are
\emph{change} (26\% of justifications, \eg{},
  happen, remain, increase)
\emph{social} (24\%, \eg{},
  vote, support, help)
\emph{cognition} (22\%, \eg{},
  think, believe, know)
and
\emph{motion} (19\%, \eg{},
  go, come, leave).
The most common noun classes are
\emph{act} (71\%, \eg{},
  election, support, deal),
\emph{communication} (57\%, \eg{},
  questions, forecast, news),
\emph{cognition} (38\%, \eg{},
  point, issue, possibility),
and
\emph{group} (38\%, \eg{},
  government, people, party).

\section{Experiments and Results}
\label{s:experiments}
We experiment with the problem of calling a question throughout its life.
The input to the problem is the question itself and forecasts (predictions and justifications),
and the output is an answer to the question aggregating all the forecasts.
The number of instances is the number of days all questions were open
(recall our dataset contains 441 questions and 96,664 forecasts submitted in 32,708 days).
We experiment both with simple baselines
and a neural network taking into account~(a)~daily forecasts
and~(b)~active forecasts submitted up to ten days prior.
Experimental results showed that considering earlier active forecasts is not beneficial.

We divide the questions into training, validation, and test subsets.
Then, we assign to each subset all the forecasts submitted throughout the life of the questions.
Note that randomly splitting forecasts would be unsound,
as forecasts for the same questions submitted on different days would end in the training, validation, and test subsets.

\begin{figure*}

\small
\centering
\includegraphics[width=\textwidth]{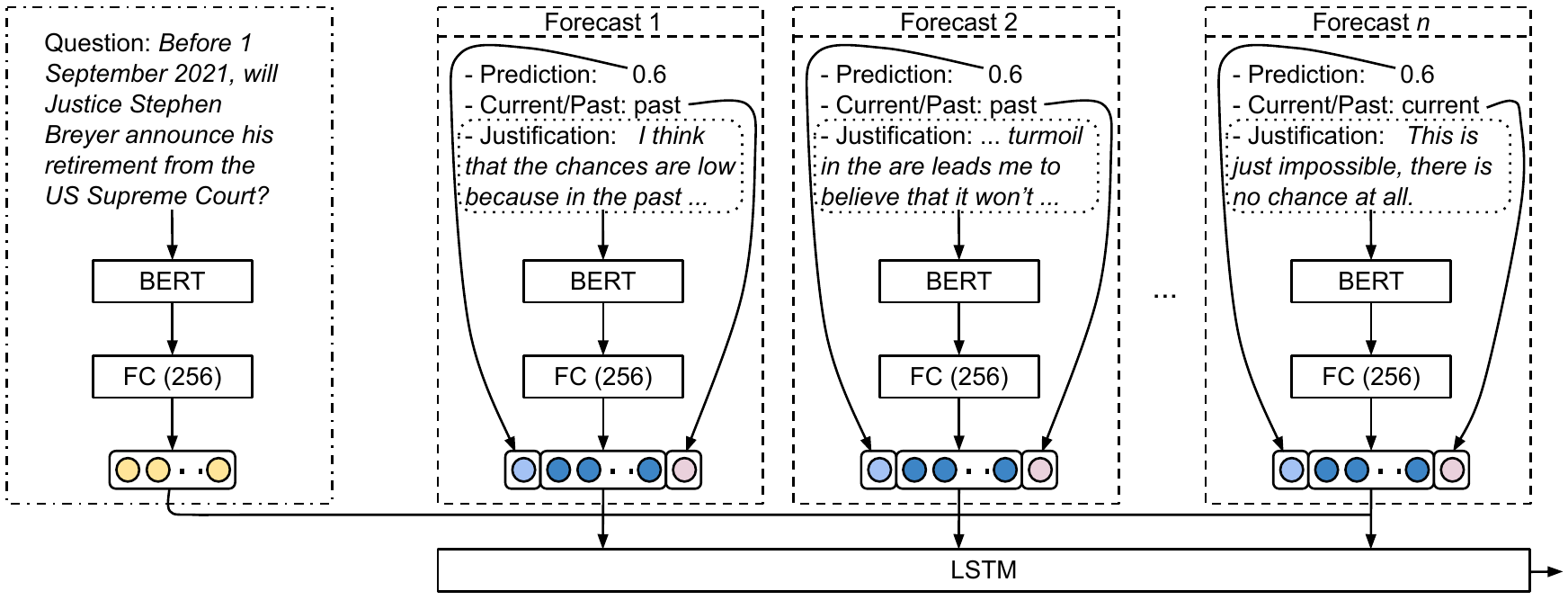}
\caption{Neural network architecture to call a question on a given day based on crowdsourced forecasts.
  The network consists of three main components:
  one for the question,
  one for each forecast (prediction + flag indicating current day or past + justification),
  and an LSTM to process the sequence of forecasts.
  We experiment with two scenarios: feeding the network the forecasts submitted on a given day (daily)
  or the last forecast by each forecaster within the ten previous days of a given day (active).
  }
\label{f:nn}
\end{figure*}

\paragraph{Baselines}
We consider two unsupervised baselines.
The \emph{majority} vote baseline calls a question based on the majority prediction in the forecasts.
The \emph{weighted} vote baseline calls a question after weighting the chances assigned to the predictions in the forecasts.
Consider these three forecasts:
99\%, 45\%, and 45\% chance the answer is \emph{yes} (thus 1\%, 55\%, and 55\% chance the answer is \emph{no}).
The majority vote baseline would output \emph{no} (2 out of~3 believe \emph{no} is more likely).
On the other hand, the weighted vote baseline would output \emph{yes}
(the weighted support for \emph{yes} is larger, 0.99 vs. 0.90).\


\subsection{Neural Network Architecture}
We experiment with the neural network architecture depicted in Figure~\ref{f:nn}.
The network has three main components:
a component to obtain a representation of the question,
a component to obtain a representation of a forecast,
and an LSTM \cite{hochreiter1997long} to process the sequence of forecasts and call the question.

We obtain the representation of a question using
BERT~\cite{devlin-etal-2019-bert}
followed by a fully connected layer with 256 neurons,
ReLU activation,
and 0.5 dropout~\cite{JMLR:v15:srivastava14a}.
We obtain the representation of a forecast concatenating three elements:
(a)~a binary flag indicating whether the forecast was submitted in the day the question is being called or in the past,
(b)~the prediction (a number ranging from 0.0 to 1.0),
and
(c)~a representation of the justification.
We obtain the representation of the justification using BERT
followed by a fully connected layer with 256 neurons, ReLU activation, and 0.5 dropout.
The LSTM has a hidden state with dimensionality 256, and takes as its input the sequence of forecasts.
During the tuning process, we discovered that it is beneficial to pass the representation of the question with each forecast
as opposed to processing forecasts independently of the question.
Therefore, we concatenate the representation of the question to each representation of a forecast prior to feeding the sequence to the LSTM.
Finally, the last hidden state of the LSTM is connected to a fully connected layer with 1 neuron and sigmoid activation to call the question.

\paragraph{Architecture Ablation}
We experiment with the full neural architecture as described above and disabling several components.
Specifically, we experiment with representing a forecast taking into account different information:
\begin{compactitem}
\item the \emph{prediction};
\item the \emph{prediction} and the representation of the \emph{question};
\item the \emph{prediction} and the representation of the \emph{justification}; and
\item the \emph{prediction}, the representation of the \emph{question}, and the representation of the \emph{justification}.
\end{compactitem}

\paragraph{Implementation and Training Details}
In order to implement the models,\footnote{Code to replicate our experiments available at 
\url{https://github.com/saketh12/forecasting_emnlp2021}}
we use the Transformers library by HuggingFace~\cite{wolf-etal-2020-transformers}
and PyTorch~\cite{NEURIPS2019_9015}.
We use binary cross-entropy loss,
gradient accumulation and mixed precision training~\cite{micikevicius2018mixed}
to alleviate the memory requirements,
the Adam optimizer~\cite{DBLP:journals/corr/KingmaB14} with learning rate 0.001,
batch size 16,
and early stopping with patience set to 3 epochs.
We tuned all the hyperparameters comparing held-out results with the validation set,
and report results with the test set.



\begin{table*}[t!]
\small
\centering
\newcommand{\sintra}{$^{\dagger}$}
\newcommand{\sinter}{$^{\ddagger}$}
\begin{tabular}{l @{\hspace{.5in}} r@{}l @{\hspace{.5in}} c@{}l c@{}l c@{}l c@{}l}
\toprule
& \multicolumn{9}{c}{days with question was open} \\ \cmidrule{2-10}
& All && Q1 && Q2 && Q3 && Q4 \\ \midrule

Using daily forecasts only \\
~~~~~~Baselines \\
~~~~~~~~~~~~Majority vote (predictions)
& 71.89 &\sintra& 64.59 &\sintra& 66.59 &\sintra& 73.26 &\sintra& 82.22 &\sintra\\
~~~~~~~~~~~~Weighted vote (predictions)
& 73.79 &\sintra& 67.79 &\sintra& 68.71 &\sintra& 74.16 &\sintra& 83.61 &\sintra\\

~~~~~~Neural network with components \ldots \\
~~~~~~~~~~~~predictions
& 77.96 &\sintra& 77.62 & & 77.93 && 78.23 && 78.61 &\sintra \\
~~~~~~~~~~~~predictions + question
& 77.61 &\sintra& 75.44 &\sinter& 76.77 &\sintra& 78.05 &\sintra& 81.56&\sintra\sinter \\
~~~~~~~~~~~~predictions + justifications
& 80.23 & \sinter & 77.87 & & 78.65 && 79.26 && 84.67&\sintra\sinter \\
~~~~~~~~~~~~predictions + question + justifications
& 79.96 &\sintra\sinter & 78.65 & & 78.11 &\sintra& 80.29 &\sintra\sinter& 83.28&\sintra\sinter \\ \midrule

Using active forecasts \\
~~~~~~Baselines \\
~~~~~~~~~~~~Majority vote (predictions)
& 77.27 & \sintra & 68.83 &\sintra& 73.92 &\sintra& 77.98 &\sintra& 87.44&\sintra \\
~~~~~~~~~~~~Weighted vote (predictions)
& 77.97 & \sintra & 72.04 &\sintra& 72.17 &\sintra& 78.53 &\sintra& 88.22&\sintra \\

~~~~~~Neural network with components \ldots \\
~~~~~~~~~~~~predictions
& 78.81 & \sintra & 77.31 &        & 78.04 && 78.53 && 81.11&\sintra  \\
~~~~~~~~~~~~predictions + question
& 79.35 & \sintra         & 76.05 &        & 78.53 &\sintra& 79.56 &\sintra& 82.94&\sintra\sinter \\
~~~~~~~~~~~~predictions + justifications
& 80.84 &         \sinter & 77.86 && 79.07 && 79.74 && 86.17&\sintra\sinter \\
~~~~~~~~~~~~predictions + question + justifications
& 81.27 &\sintra\sinter& 78.71 &\sinter& 79.81 &\sintra\sinter& 81.56 &\sintra\sinter& 84.67&\sintra\sinter \\

\bottomrule

\end{tabular}
\caption{Results with the test questions (Accuracy, \ie{}, the average percentage of days a model calls a question correctly).
  We provide results with \emph{All} days a question was open and four quartiles
  (Q1: first 25\% of days, Q2: 25--50\%, Q3: 50--75\%, and Q4: last 25\% of days).
  We calculate statistical significance (McNemar's test \cite{mcnemar1947note} with $p < 0.05$) between 
  (a)~each model using daily or active forecasts
  (all models obtain significantly better results using the active forecasts except the neural network with the \emph{predictions + justifications} component, indicated with \sintra)
  and
  (b)~the neural network trained with the \emph{predictions} component and the networks trained with the additional components    
  (adding the \emph{justification} and both the \emph{question and justification} yields significantly better results using  daily or active forecasts, indicated with \sinter).}
\label{t:results}
\end{table*}

\subsection{Quantitative Results}

Table \ref{t:results} presents the results.
The evaluation metrics is accuracy (\ie{}, average percentage of days a model calls a question correctly throughout the life of the question).
We report results for all days~(column~2)
and
the four quartiles~(columns~3--6).

Despite their simplicity, the baselines obtain good results~(71.89 and 73.79 using daily and active forecasts),
showing that aggregating the predictions submitted by forecasters without regard to the question or justifications is a competitive approach.
As we shall see, however,
the full neural network obtains statistically significant better results (79.96 and 81.27 using daily and active forecasts).

\paragraph{Using Daily or Active Forecasts}
Taking into account active forecasts instead of only those submitted on the day the model is calling the question~(daily forecasts)
is beneficial across both baselines and all neural networks except the one using only \emph{predictions + justification}.
The differences in accuracy
are larger with the baselines (daily: 71.89 vs. 77.27; active: 73.79 vs. 77.97)
than with the neural networks.
We note, however, that the differences are statistically significant evaluating with all days and all quartiles except Q1~(indicated with~$^{\dagger}$ in Table \ref{t:results}, McNemar's test \cite{mcnemar1947note} with $p < 0.05$).
\textbf{We conclude that using active forecasts is beneficial}
and focus the remaining of the discussion on these results.

\begin{table*}
\small
\centering
\newcommand{\sintra}{$^{\dagger}$}
\newcommand{\sinter}{$^{\ddagger}$}
\begin{tabular}{l @{\hspace{.5in}} r@{}l @{\hspace{.5in}} c@{}l c@{}l c@{}l c@{}l}
\toprule
& \multicolumn{9}{c}{question difficulty (according to best baseline)} \\ \cmidrule{2-10}
& All && Q1 && Q2 && Q3 && Q4 \\ \midrule



Using active forecasts \\
~~~~~~Weighted vote baseline (predictions)
&77.97&  & 99.40 && 99.55 && 86.01 && 29.30 \\

~~~~~~Neural network with components \ldots \\
~~~~~~~~~~~~predictions + question
&79.35&\sintra  & 94.58 &\sintra\sinter&  88.01 &\sintra\sinter& 78.04 &\sintra\sinter& 58.73&\sintra\sinter \\
~~~~~~~~~~~~predictions + justifications
&80.84&\sintra\sinter  & 95.71 &\sintra\sinter& 93.18 &\sintra\sinter& 79.99 &\sintra\sinter&  57.05&\sintra\sinter \\
~~~~~~~~~~~~predictions + question + justifications
&81.27&\sintra\sinter  & 94.17 &\sintra\sinter& 90.11 &\sintra\sinter& 78.67 &\sintra\sinter& 64.41&\sintra\sinter \\

\bottomrule

\end{tabular}

\caption{Results with the test questions (Accuracy, \ie{}, average percentage of days a question is called correctly).
  We provide results with \emph{All} questions and depending on the question difficulty as measured by the results obtained with the best baseline
  (Q1: easiest 25\%; Q2: 25--50\%, Q3: 50--75\%, and Q4: hardest 25\%).
  We calculate statistical significance (McNemar's test \cite{mcnemar1947note} with $p < 0.05$) between 
  (a)~the weighted vote baseline and each neural network (indicated with \sintra), and
  (b)~the neural network trained with the \emph{predictions} component (not shown)
  and the networks trained with the additional components (indicated with \sinter).
}
\label{t:extra_results}
\end{table*}

\paragraph{Encoding Questions and Justifications}
The neural network that uses only the \emph{prediction} to represent a forecast outperforms both baselines~(78.81 vs. 77.27 and 77.97).
More interestingly, incorporating into the representation of the forecast
the question,
the justification,
or both brings improvements (79.35, 80.84, and 81.27).
All but the results with \emph{predictions + justifications} are statistically significant with respect to using only \emph{predictions}.
\textbf{We conclude that calling a question benefits from incorporating into the model the question and the justifications submitted by forecasters}.

\paragraph{Calling Questions Throughout their Life}
We now move beyond accuracies calculated using all days throughout the life of a question
and examine detailed results per quartile.
More specifically,
we divide the days into four quartiles.
The last four columns in Table \ref{t:results} show that while using active forecasts is beneficial across all four quartiles~(with both baselines and all networks),
the neural networks---perhaps surprisingly---outperform the baselines only in the first three quartiles.
In fact, the neural networks obtain statistically significant worse results than any of the baselines in the last quartile
(84.67 vs. 87.44 and 88.22; -3.2\% and -4.0\%).
\textbf{We conclude that modeling questions and justifications is overall useful,
although it is detrimental towards the end of the life of a question}.
The justification for this empirical fact is that the crowd gets wiser towards the end of the life of a question---as more evidence to make the correct prediction presumably becomes available, and more forecasters submit forecasts.
Our model does not take into account which day is calling a question in (within the life of the question).
We reserve to future work incorporating temporal information to better aggregate forecasts.


\paragraph{Calling Questions Based on their Difficulty}
We finish the quantitative experiments with results depending on the difficulty of the questions.
To this end,
we sort questions by their difficulty based on how many days the majority or weighted vote baselines (whichever makes the least mistakes) calls the questions wrong.
These experiments shed light into how many questions benefit from the neural networks 
that take into account the question and justifications.
We note, however, that it is impossible to calculate question difficulty during the life of the question, so these experiments are not realistic before a question closes (and the correct answer is known).
After all, forecasts are about predicting the future, and it is only challenging to do so while the correct answer is unknown.

Table \ref{t:extra_results} shows the results with selected models depending on question difficulty.
We observe that the weighted vote baseline calls 75\% questions more reliably than the neural network.
Indeed, the baseline obtains 99.40, 99.55, 86.01 and 29.30 accuracy in each quartile of difficulty,
while the best network obtains
95.71 (-3.7\%),
93.18 (-6.4\%),
79.99 (-7.0\%), and
64.41 (+119.8\%).
In other words,
the majority of questions (75\% easiest questions) obtain worse results with the best neural network (-3.7--7.0\%),
but a substantial amount (25\% hardest questions) are called correctly more than twice as often (+119.8\%).
The benefits with the hardest questions compensate the drawbacks with the easiest questions.
As stated earlier, overall the full neural network obtains significantly better results than the baselines~(81.27 vs. 77.27 and 77.97).
\textbf{We conclude that learning how to aggregate crowdsourced forecasts},
and specifically taking into account the question and justifications,
\textbf{is the most beneficial with the hardest questions.}
\section{Qualitative Analysis}
\label{s:qualitative_analysis}

\begin{table*}[h]
\small
\centering
\newcommand\cw{\hspace{.85in}}

\begin{tabular}{l cc}

\toprule
& \multicolumn{2}{c}{Questions \ldots} \\ \cmidrule{2-3}
& called wrong $\ge 1$ day & always called correct \\ \midrule

\# days open           & 69.4 & 81.7 \\
\# forecasts available & 31.0 & 26.7 \\
\% incorrect forecasts & 49.7 & 16.6 \\ \bottomrule \addlinespace

\toprule
& \multicolumn{2}{c}{Justifications submitted with \ldots} \\ \cmidrule{2-3}
& wrong predictions & correct predictions \\ \midrule

\% short ($<$ 20 tokens) &
78.0 & 65.0 \\
\% with references to previous forecasts &
31.5 & 16.0 \\
\% without a logical argument &
62.5 & 47.5 \\
\% with generic arguments &
16.0 & 14.5 \\
\% with poor grammar or spelling, non-English &
24.5 & 14.5 \\ \bottomrule

\end{tabular}

\caption{Characterizations of questions and justifications based on the predictions obtained with the best model
(NN with \emph{predictions + question + justification} trained with active forecasts, Table \ref{t:results}).
The top block characterizes all questions in the test set (88 questions) depending on whether the model calls the question wrong in one day.
The bottom block characterizes 400 random justifications from days that the model calls a question wrong (200 written justifications submitted with correct and wrong forecasts each).}
\label{t:error_analysis}
\end{table*}

In this section, we present insights into
(a)~what makes questions harder to forecast
and
(b)~characteristics of justification submitted with wrong and correct predictions (Table \ref{t:error_analysis}).

\paragraph{Questions}
We looked at three characteristics of the 88 questions in the test set depending on whether the best model
(bottom row in Table \ref{t:results})
calls the question at least one day wrong (it does so with 36 out of 88 questions).
Surprisingly, we found that questions that are called correct in all days
have a longer life (81.7 vs. 69.4 days) and less active forecasts per day (26.7 vs. 31.0).
As one would expect,
our best model makes mistakes with the same questions that forecasters struggle with.

\paragraph{Justifications.}
We manually analyzed 200 justifications submitted with wrong and correct predictions (400 in total).
Specifically, we looked at predictions submitted on days that our best model makes a mistake calling the corresponding question.
Here are the observations we identified:
\begin{compactitem}

\item We found that 78\% of wrong predictions were submitted with \emph{short justifications} (less than 20 tokens),
  while 65\% of correct predictions were.
  This observation corroborates that longer user-generated text has higher quality~\cite{beygelzimer2015yahoo}.
\\Example: \emph{Software isn't good enough yet},
  submitted to question \emph{Will Google's AlphaGo beat world champion Lee Sedol in the five-game Go match planned for March 2016?}

\item While relatively few forecasts \emph{refer to previous forecasts} (by the same or other forecasters, or the current forecast by the crowd),
  we observe that justifications for wrong predictions do almost twice as often (31.5\% vs. 16.0\%).
\\Example: \emph{Returning to initial forecast}.

\item \emph{Lack of logical arguments} is common in the justifications we work with.
This is true regardless of whether the predictions they were submitted with are wrong or correct. 
  We found, however, that not having a logical argument is more common with wrong predictions~(62.5\% vs. 47.5\%).
\\Example: \emph{I guess Greek head of state does not count, but we are getting close},
submitted to question \emph{Will Iran host a head of state or government from one of the G7 countries on an official visit before 1 July 2016?}

\item Surprisingly, justifications with \emph{generic arguments} are not a clear indicator of wrong or correct predictions (16.0\% vs. 14.5\%).
\\Example: \emph{It seems to be pretty much decided, unless something completely out of the blue happens}.

\item \emph{Poor grammar and spelling or non-English} are rare, but much more common in justification of wrong predictions (24.5\% vs. 14.5\%).
\\Example: \emph{For reference y'all} and \emph{Wenn Trump den Kurs beibehAlt}.

\end{compactitem}

\section{Conclusions}
\label{s:conclusions}

Forecasting is the process of predicting future events.
Government and industry alike are interested in forecasting because
it affords them the capability to anticipate and address potential challenges to come.
In this paper, we work with questions across the political, economic, and social spectrum published in the Good Judgment Open website,
and forecasts submitted by the crowd without special training.
Each forecast consists of a prediction and a justification in natural language.

We have shown that aggregating the weighted predictions of forecasters is a robust baseline to call a question through its life.
Models that take into account both the question and justifications, however,
obtain significantly better results when calling a question in the first three quartiles of its life.
Crucially, our models do not profile forecasters or use any information about who submitted which forecast.
The work presented here
opens the door to assessing the credibility of anonymous forecasts in order to come up with aggregation strategies that are robust without tracking forecasters.

\section*{Acknowledgments}
\label{s:acknowledgments}

We gratefully acknowledge the support of NVIDIA Corporation with the donation of the Titan Xp GPU used for this research.
The results presented in this paper were also obtained using the Chameleon testbed supported by the National Science Foundation \cite{keahey2020lessons}.

\bibliographystyle{emnlp2021-latex/acl_natbib}
\bibliography{refs}


\end{document}